\documentclass[conference]{IEEEtran}
\IEEEoverridecommandlockouts

\usepackage{cite}                       
\usepackage{hyperref}
\usepackage{url}

\usepackage{amsmath,amssymb,amsfonts}   
\usepackage{algorithmic}                

\usepackage{graphicx}                   
\usepackage{subcaption}
\usepackage{booktabs}
\usepackage{multirow}
\usepackage{array}
\usepackage{adjustbox}
\usepackage{placeins}
\usepackage{xcolor}
\usepackage{colortbl}
\usepackage{tikz}
\usetikzlibrary{shapes.geometric, arrows.meta, positioning}

\usepackage{tikz}                       
\usepackage{pgfplots}
\usepackage{tcolorbox}                  
\pgfplotsset{compat=1.18}

\usepackage{textcomp}                   
\usepackage{xcolor}
\usepackage{comment}
\usepackage{pifont}

\usepackage{orcidlink}                  

\usetikzlibrary{shapes.geometric, arrows.meta, positioning, fit, backgrounds, decorations.pathreplacing}  

\begin{document}
\raggedbottom
\title{Agentic AI for Commercial Insurance Underwriting with Adversarial Self-Critique}

\author{\IEEEauthorblockN{1\textsuperscript{st} Joyjit Roy
}
\IEEEauthorblockA{\textit{IEEE Senior Member} \\
Austin, Texas, USA \\
joyjit.roy.tech@gmail.com}
\and
\IEEEauthorblockN{2\textsuperscript{nd} Samaresh Kumar Singh
}
\IEEEauthorblockA{\textit{IEEE Senior Member} \\
Leander, Texas, USA \\
ssam3003@gmail.com}
}
\maketitle

\begin{abstract}
Commercial insurance underwriting is a labor-intensive process that requires manual review of extensive documentation to assess risk and determine policy pricing. While AI offers substantial efficiency improvements, existing solutions lack comprehensive reasoning and internal mechanisms to ensure reliability in regulated, high-stakes environments. Full automation remains impractical and inadvisable when human judgment and accountability are critical. This study presents a decision-negative, human-in-the-loop agentic system that incorporates an adversarial self-critique mechanism as a bounded safety architecture for regulated underwriting workflows. In this system, a critic agent challenges the primary agent's conclusions prior to submitting recommendations to human reviewers. This internal system of checks and balances addresses a critical gap in AI safety for regulated workflows. Additionally, the research develops a formal taxonomy of failure modes to characterize potential errors by decision-negative agents. This taxonomy provides a structured framework for risk identification and management in high-stakes applications. Experimental evaluation using 500 expert-validated underwriting cases demonstrates that the adversarial critique mechanism reduces AI hallucination rates from 11.3\% to 3.8\% and increases decision accuracy from 92\% to 96\%. At the same time, the framework enforces strict human authority over all binding decisions by design. These findings indicate that adversarial self-critique supports safer AI deployment in regulated domains and offers a model for responsible integration where human oversight is indispensable.
\end{abstract}

\begin{IEEEkeywords}
Agentic AI, 
Human-in-the-Loop, 
Insurance Underwriting, 
Large Language Models (LLMs), 
Adversarial Self-Critique, 
Regulated AI Systems
\end{IEEEkeywords}
\section{Introduction}
\label{sec:introduction}

Commercial insurance underwriting typically takes days to weeks per submission as underwriters manually review extensive documentation. AI has strong potential to accelerate this process. For standard policies, average decision time dropped from 3-5 days to about 12 minutes while maintaining 99.3\% accuracy \cite{biztech2025}. Even in complex cases, AI reduced processing time by 31\% and improved accuracy by 43\% \cite{deloitte2025}.

However, full automation is neither feasible nor advisable in this regulated environment. Regulatory bodies require human oversight for high-risk AI systems \cite{euaiact2024, fenwick2025}. AI failure modes pose real risks. Without mitigation, LLMs hallucinate frequently. Rates reach 15\% in finance-related tasks \cite{huang2023hallucination}. Current insurance AI solutions either handle narrow tasks or operate as black boxes without internal audit mechanisms.

This work introduces an adversarial self-critique mechanism where a critic agent challenges the primary agent's conclusions before human review. The work also develops a formal failure-mode taxonomy for decision-negative agents. Experimental evaluation on 500 expert-validated cases shows 96\% decision accuracy versus 92\% without critique. Hallucination rates dropped from 11.3\% to 3.8\%. The system delivers 4-6× efficiency gains while preserving human final authority.

\section{Related Work and Research Gap}
\label{sec:related_work}

\subsection{Agentic AI Systems}

Recent agentic AI frameworks such as AutoGPT, BabyAGI, and ReAct enable autonomous agents to plan actions and use tools without constant human prompting \cite{yang2023autogpt, yao2023react}. These systems leverage LLMs to interweave reasoning and acting in autonomous loops. However, practical use in high-stakes domains like underwriting has revealed significant limitations. Without explicit constraints, autonomous agents may pursue objectives in unpredictable ways, which is unacceptable in regulated environments where compliance is critical.

No off-the-shelf framework provides the fine-grained control 
and oversight required for regulated insurance workflows. The approach presented here applies agentic AI within a bounded environment, enforcing strict constraints such as read-only tool use and no final decision authority. It also incorporates an internal critique loop to proactively evaluate its actions before surfacing recommendations to human reviewers.

\subsection{AI in Insurance Underwriting}

Insurance firms have begun adopting AI in underwriting through ML risk models and intelligent document processing. Initial deployments focus on submission triage, data extraction, and fraud detection \cite{shift2024}. These specialized tools have increased operational throughput. Some insurers report underwriting time reduced by roughly 30\% with accuracy gains of approximately 40\% \cite{decerto2025}. However, full decision automation in commercial underwriting remains rare.

Despite these gains, significant barriers remain. Most underwriting leaders identify ineffective legacy systems, poor data quality, and limited data accessibility as impediments \cite{accenture2025}. AI adoption has been limited and fragmented, automating only segments of the process rather than entire workflows. Underwriters and regulators remain cautious due to the opacity of AI decision-making and potential compliance risks.

\subsection{Human-in-the-Loop AI}

Human-in-the-loop designs have proven effective across multiple high-stakes domains. In radiology, FDA-cleared AI systems flag anomalies for mandatory physician review before any clinical use action \cite{topol2019}. In legal AI, professional responsibility regulations require mandatory attorney review before AI-assisted outputs affect client matters \cite{euaiact2024}. In ecommerce, AI-powered catalog automation uses human validation for high-risk attributes. This achieves 99.6\% upload accuracy and reduces seller rework by 61\% \cite{transformsolutions2025}.

The approach here aligns with an augmented model. Routine cases are handled largely by AI. Cases with uncertainty or policy exceptions get escalated to human judgment. Underwriters can override any AI recommendation. This follows the National Association of Insurance Commissioners' principle that AI should inform, not replace, human decisions \cite{fenwick2025}.

\subsection{Adversarial Self-Critique in AI}

Recent research in AI safety and reliability explores how AI models can critique or reflect on their outputs to improve quality. Techniques like Anthropic's Constitutional AI use an AI-generated set of principles to enable the model to self-evaluate and revise its answers, reducing harmful or inaccurate outputs \cite{bai2022constitutional}. Another line of work involves multi-agent debate, where one agent produces an answer and another critiques it, leading to more refined responses.

Zheng et al. introduced Critic Chain-of-Thought, where an LLM generates a reasoning chain and a separate critic chain evaluates it \cite{zheng2025critic}. This yielded significant accuracy gains on math word problems by filtering out invalid intermediate steps. Kamoi et al. surveyed self-correction strategies, noting that LLMs rarely correct their mistakes without reliable feedback and that simple prompting is often inadequate \cite{kamoi2024selfcorrection}. The takeaway is that structured critique, whether external or internal, can strengthen a model's reasoning ability. 

\subsection{Research Gap}

Table~\ref{tab:relatedwork} summarizes the landscape of related work and highlights the gaps this research addresses. 

\begin{table}[h]
\centering
\caption{Comparison of AI Approaches in Regulated Domains}
\label{tab:relatedwork}
\begin{tabular}{lccccc}
\hline
\textbf{System/Approach} & \textbf{E2E} & \textbf{Int.} & \textbf{HITL} & \textbf{Bounded} & \textbf{Fail.} \\
 & \textbf{Reas.} & \textbf{Crit.} &  & \textbf{Auth.} & \textbf{Tax.} \\
\hline
AutoGPT/ReAct \cite{yang2023autogpt,yao2023react} & \checkmark & \texttimes & \texttimes & \texttimes & \texttimes \\
Insurance AI Tools \cite{shift2024} & \texttimes & \texttimes & \checkmark & \checkmark & \texttimes \\
Constitutional AI \cite{bai2022constitutional} & \checkmark & \checkmark & \texttimes & \texttimes & \texttimes \\
Critic CoT \cite{zheng2025critic} & \checkmark & \checkmark & \texttimes & \texttimes & \texttimes \\
\hline
\textbf{This Work} & \checkmark & \checkmark & \checkmark & \checkmark & \checkmark \\
\hline
\end{tabular}
\footnotesize{\textit{Note:} E2E Reas. = End-to-End Reasoning; Int. Crit. = Internal Critique; HITL = Human-in-the-Loop; Bounded Auth. = Bounded Authority; Fail. Tax. = Failure Taxonomy}
\end{table}

Prior work has not systematically studied adversarial self-critique in bounded agentic systems under strict authority constraints in regulated underwriting. No research has developed a failure-mode taxonomy for decision-negative agents in this context. Existing insurance AI solutions handle narrow tasks or operate as black-box models without internal critics. None have documented how AI can proactively test itself under human oversight. This work introduces such a mechanism and characterizes its failure modes through experimental evaluation.
\section{System Architecture}
\label{sec:architecture}

\subsection{Overview}

\begin{figure}[htbp]
\centering
\makebox[\columnwidth][c]{%
  \includegraphics[width=0.85\columnwidth]{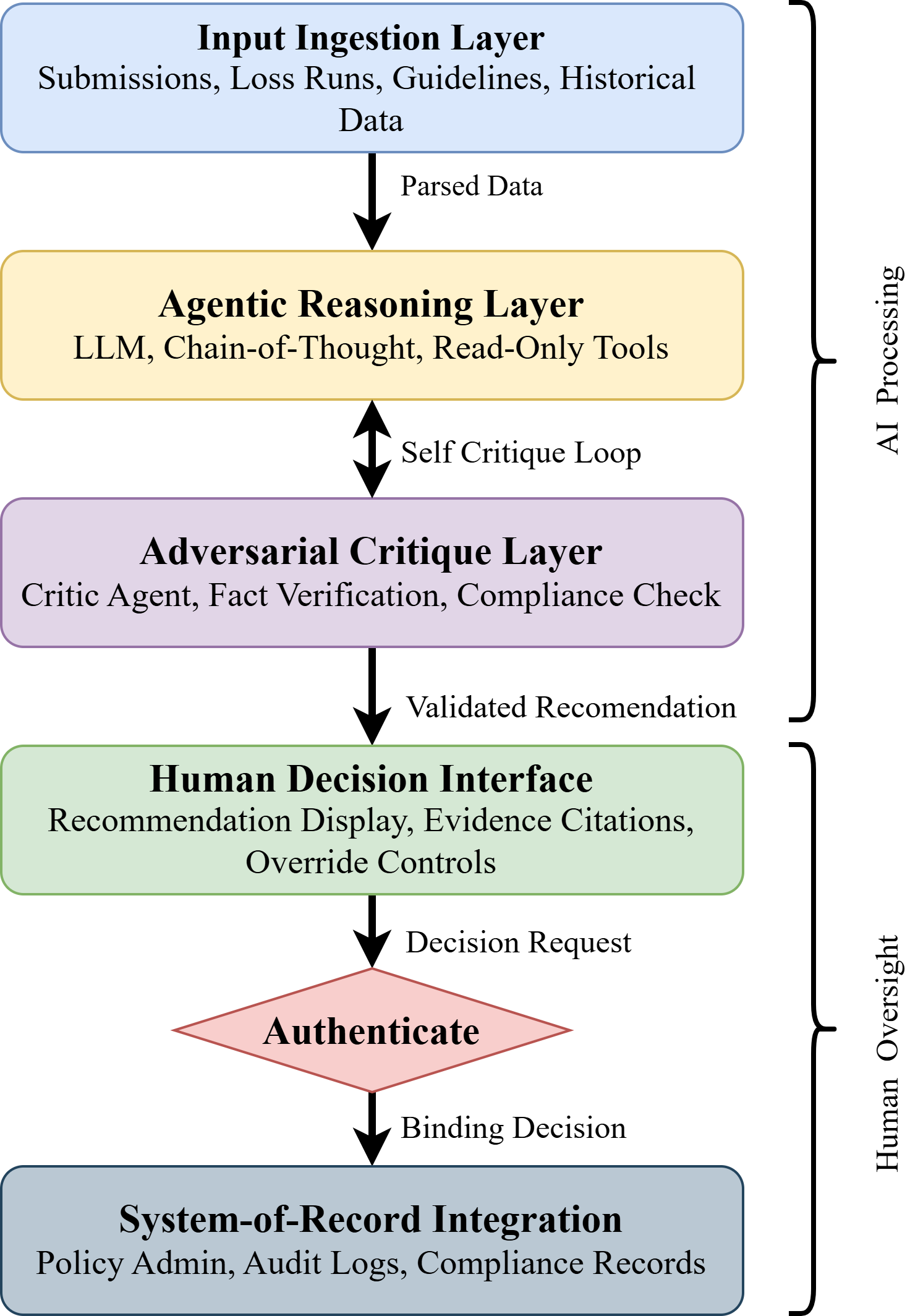}
}
\caption{State machine workflow with guard conditions. The self-critique cycle allows one iteration before proceeding to the decision. Human authorization checkpoint ensures no autonomous binding.}
\label{fig:architecture}
\end{figure}

The system architecture separates AI processing from human authority through 5 layers (Figure~\ref{fig:architecture}). The first 3 layers handle automation. \textbf{Input ingestion} consolidates submission data and guidelines. \textbf{Agentic reasoning} performs risk analysis using chain-of-thought decomposition. \textbf{Adversarial critique} validates the primary agent's conclusions. The final two layers ensure human control. A \textbf{human decision} interface presents recommendations with full traceability. \textbf{System-of-record} integration logs outcomes only after human authorization. The AI remains decision-negative and requires human approval before any binding action.

\subsection{Workflow}

When a submission arrives, the Input Layer gathers documents, inspection reports, financial statements, underwriting guidelines, and historical account data. This feeds the Agentic Reasoning Layer, which uses Claude Sonnet 4.5 for its extended context window of 200,000 tokens \cite{anthropic2025claude}. The agent processes inputs through chain-of-thought reasoning, breaking underwriting into discrete steps. These include extracting risk factors, verifying compliance with guidelines, and computing preliminary premiums. Additionally, the agent may invoke read-only tools for database queries, location-risk verification, or vector search in the underwriting manual.

Once the primary agent formulates a draft decision, the Adversarial Critique Layer activates. The critic agent receives the draft output and reasoning chain, tasked to identify errors, unsupported assumptions, or guideline violations. The critic is prompted to assume the role of a skeptical internal reviewer who cross-checks every stated fact against input data and guidelines, lists discrepancies, and may inject hypothetical scenarios to test decision robustness.

The primary agent then revises its decision, addressing the critic's points. One full critique-revision cycle is allowed to balance thoroughness with efficiency. This self-critique acts as an internal check, where the agent must convince an adversarial version of itself before reaching human review.

The refined recommendation is presented to the human underwriter via the Decision Interface, showing key evidence, reasoning, and citations to source documents. All data is traceable from origin to decision. The underwriter can question the AI's rationale through a chat interface. The underwriter makes the binding decision by accepting, modifying, or overriding the AI's recommendation. The system records the final decision, AI analysis, and human actions in an audit log.

\subsection{Authority Boundaries}

\begin{figure*}[htbp]
\centering
\includegraphics[width=\textwidth]{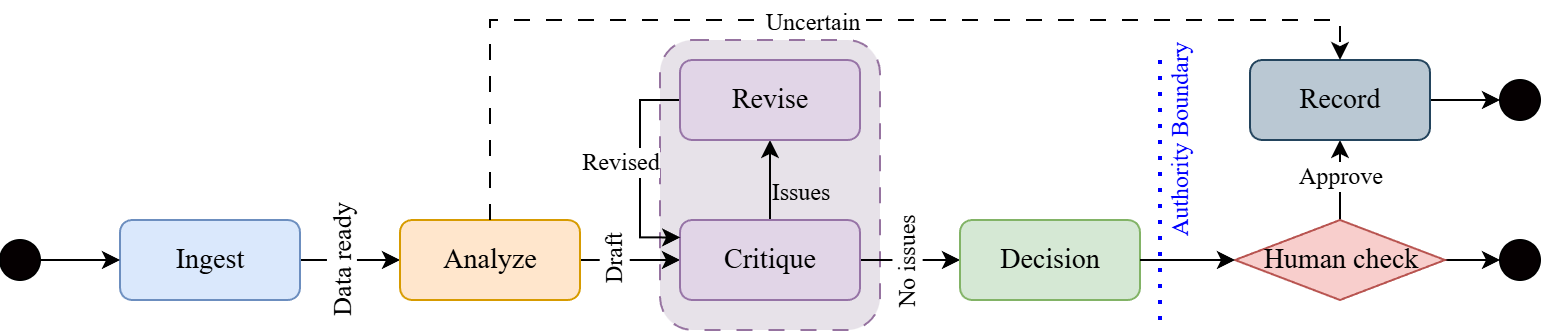}
\caption{State machine workflow with guard conditions. The self-critique cycle allows one iteration before proceeding to the decision. Human authorization checkpoint ensures no autonomous binding. The dashed path shows graceful degradation for uncertain cases.}
\label{fig:statemachine}
\end{figure*}

The workflow is structured as a state machine with guard conditions that enforce human-in-the-loop constraints (Figure~\ref{fig:statemachine}). Transitions occur sequentially from Ingest to Analyze, Critique, Revise, and Decision, with the critic initiating revision when issues are identified. The Decision state produces only a recommendation, never a binding outcome. A human authorization checkpoint gates the transition to the Record state, where decisions are finalized. In cases of high uncertainty, the system escalates immediately to human review through a graceful degradation path.

Several guardrails restrict the agent's operation. Tool usage is limited to read-only data retrieval APIs. The output format is validated against a schema that requires structured fields, including Recommendation, Supporting Facts, and Flags, and excludes any field for executing binding actions. Prompts explicitly instruct the agent to follow guidelines and not fabricate information. Fail-safes flag unsupported scenarios for human review rather than attempting auto-resolution.

\subsection{Implementation Details}

Both agent roles use Claude Sonnet 4.5, the strongest model for complex agents as of 2025 \cite{anthropic2025claude}. Agent and critic prompts are orchestrated by a controller using a state machine framework analogous to LangGraph. The controller manages state transitions and triggers a human-in-the-loop escalation if outputs fail to converge.

\begin{figure*}[htbp]
\centering
\includegraphics[width=\textwidth]{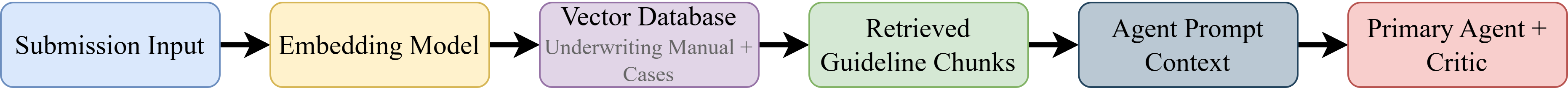}
\caption{RAG retrieval pipeline. Submission text is embedded and matched to the carrier's vector database. Retrieved guideline chunks ground both the primary agent and critic in the approved policy instead of parametric memory.}
\label{fig:rag}
\end{figure*}

A vector database stores the carrier's underwriting manual and historical cases. Embedding-based retrieval grounds recommendations in exact guideline text rather than in parametric memory, reducing hallucinations. Figure~\ref{fig:rag} illustrates the retrieval pipeline. After the underwriter accepts a recommendation, the audit log lets regulators trace each input from its origin to the final decision.

\subsection{Prompt Design}

Both agent roles follow structured prompt templates that define their scope, reasoning format, and output constraints. Representative skeletons are provided below. Carrier-specific guideline excerpts injected via retrieval augmentation are omitted because they are proprietary to the deploying insurer.

\vspace{4pt}
\begin{tcolorbox}[
  title=Primary Agent Prompt (abbreviated),
  fonttitle=\bfseries\footnotesize, fontupper=\footnotesize,
  colback=gray!12, colframe=black!60, coltitle=white,
  boxrule=0.6pt, arc=2pt, left=4pt, right=4pt, top=4pt, bottom=4pt
]
You are a commercial insurance underwriting assistant. Analyze the submission below and produce a structured risk recommendation. Use only the provided documents and the retrieved guideline sections. Do not fabricate facts. Do not issue binding decisions. Flag missing or ambiguous information for human review.

\vspace{4pt}
Reason step by step. For each risk factor, cite the source document or guideline section.

\vspace{4pt}
\textbf{Output schema:} \texttt{Recommendation} (Bind $\mid$ Decline $\mid$ Refer); \texttt{Supporting\_Facts} (bulleted, source-cited); \texttt{Flags} (unresolved issues or uncertainties).
\end{tcolorbox}

\vspace{2pt}
\begin{tcolorbox}[
  title=Critic Agent Prompt (abbreviated),
  fonttitle=\bfseries\footnotesize, fontupper=\footnotesize,
  colback=gray!12, colframe=black!60, coltitle=white,
  boxrule=0.6pt, arc=2pt, left=4pt, right=4pt, top=4pt, bottom=4pt
]
You are a skeptical internal reviewer. A draft underwriting recommendation is provided below. Challenge it. Check every stated fact against the submission. Identify unsupported assumptions, guideline violations, and logical inconsistencies. Inject at least one hypothetical stress scenario to test decision robustness.

\vspace{4pt}
Do not approve the recommendation unless all material claims are verifiable. Output a numbered list of issues with severity tags. If no issues exist, state that explicitly with justification.

\vspace{4pt}
\textbf{Output schema:} \texttt{Issues} (numbered, severity-tagged); \texttt{Verdict} (Approved $\mid$ Requires\_Revision).
\end{tcolorbox}

\vspace{4pt}
The output schema for both roles excludes any field that could trigger a binding action, serving as a programmatic guardrail against authority boundary violations.

\section{Dataset and Experimental Setup}
\label{sec:experimental_setup}

\subsection{Dataset}

The evaluation uses the Snorkel AI Multi-Turn Insurance Underwriting dataset \cite{snorkelai2025}. This is an expert-validated collection of approximately 1,000 commercial underwriting conversations. Each conversation contains 10-20 turns covering appetite checks, coverage reviews, policy comparisons, risk assessments, premium calculations, and compliance verifications.

A total of 500 cases were selected for this study and stratified by complexity:\\
(1) 100 simple cases (straightforward renewals with minimal documentation),\\
(2) 250 medium cases (standard submissions with typical risk profiles), and \\
(3) 150 complex cases (multi-coverage policies, unusual risks, or extensive documentation requiring nuanced judgment).\\ 
This stratification enabled testing across both routine and challenging scenarios.

\subsection{Ground Truth and Baselines}

Each case includes expert underwriter decisions and rationales as ground truth. Ground truth was established through review by experienced commercial insurance underwriters using consensus-based validation. In cases of disagreement, decisions were adjudicated by a senior reviewer to arrive at a final reference outcome. Expert rationales were used solely for evaluation and error analysis and were not used to train, fine-tune, or adapt the AI system. This process ensures that reported performance metrics reflect expert judgment rather than model alignment. Three configurations were evaluated:\\
(1) Human-Only Manual Process represents traditional workflow,\\
(2) LLM Agent without Critique, and\\
(3) Full System with Agent and Adversarial Critic.\\
This comparison isolates the impact of the self-critique mechanism.

\subsection{Configuration}

Claude Sonnet 4.5 was deployed via API without fine-tuning on insurance data. Few-shot prompting and retrieval augmentation provided domain specialization. A fabricated mini-submission with step-by-step reasoning and critique was prepended to demonstrate the expected format. Relevant guideline excerpts from the vector database were fed into each prompt. The temperature was set to 0.2 for the agent and 0.0 for the critic to maximize consistency.

\subsection{Statistical Analysis}
All proportions are reported with 95\% confidence intervals 
using the Wilson score method \cite{wilson1927}. McNemar's 
test assessed the statistical significance of performance differences 
between Agent Only and Agent+Critic configurations ($\alpha = 0.05$) 
\cite{mcnemar1947}. The sample size (n=500) provides 80\% power 
to detect a 4\% difference in decision accuracy.

\subsection{Procedure}

Each case was evaluated under all 3 configurations. For manual scenarios, underwriter time and baseline performance were measured. For AI scenarios, metrics included recommended decisions, errors relative to ground truth, and processing time. Domain experts reviewed each AI output, provided qualitative feedback, and classified errors for failure analysis.

Stress tests probed system robustness using adversarially perturbed cases with inconsistent or incomplete data. They simulated real-world noise such as missing fields and contradictory documents. They also included out-of-distribution cases from unseen lines of business. All experiments ran between November and December 2025 using the same model version with a reset state between cases. Statistical significance was assessed using McNemar's test where applicable.
\section{Evaluation Metrics}
\label{sec:evaluation_metrics}

Performance was evaluated across three categories. \textbf{Quality metrics} included decision accuracy (matching ground-truth decisions), risk factor recall (identifying key risks), guideline compliance rate, hallucination rate (ungrounded claims), and source traceability. \textbf{Critic-specific metrics} measured catch rate (true issues flagged), false positive rate, and correction success rate. \textbf{Efficiency metrics} tracked processing time, human effort reduction, and underwriter override rate.

Four test matrices assessed system behavior: ablation (isolating critic impact), case difficulty (simple/medium/complex), robustness (adversarial scenarios), and authority boundaries (verifying decision-negative constraints).
\section{Experimental Results}
\label{sec:results}

\subsection{Overall Performance}

Table~\ref{tab:performance} presents a performance comparison across the 3 systems. The agent with critic achieved 96\% decision accuracy (95\% CI: 94.2--97.8\%) compared to 92\% (95\% CI: 89.4--94.6\%) for the agent-only system and 100\% for the manual baseline. McNemar's test confirmed statistically significant improvement (p = 0.003).

\begin{table}[htbp]
\caption{System Performance Comparison}
\begin{center}
\label{tab:performance}
\scriptsize 
\begin{tabular}{@{}p{2cm}p{0.6cm}p{1.8cm}p{1.8cm}p{0.5cm}@{}}
\toprule
\textbf{Metric} & \textbf{Manual} & \textbf{Agent Only} & \textbf{Agent+Critic} & \textbf{p-value} \\
\midrule
Decision Accuracy & 100\%* & 92\% (89.4-94.6) & 96\% (94.2-97.8) & 0.003** \\
\arrayrulecolor{lightgray}\hline\arrayrulecolor{black}
Hallucination Rate & N/A & 11.3\% (8.7-13.9) & 3.8\% (2.3-5.3) & $<$0.001*** \\
\arrayrulecolor{lightgray}\hline\arrayrulecolor{black}
Evidence Completeness & 62\% & 87\% (84.0-90.0) & 91\% (88.4-93.6) & 0.042* \\
\arrayrulecolor{lightgray}\hline\arrayrulecolor{black}
Contradiction Detection & N/A & 76\% (72.3-79.7) & 89\% (86.2-91.8) & $<$0.001*** \\
\arrayrulecolor{lightgray}\hline\arrayrulecolor{black}
Source Traceability & N/A & 81\% (77.6-84.4) & 96\% (94.2-97.8) & $<$0.001*** \\
\arrayrulecolor{lightgray}\hline\arrayrulecolor{black}
Guideline Compliance & $\sim$100\% & 95\% (92.8-97.2) & 98.5\% (97.1-99.9) & 0.008** \\
\arrayrulecolor{lightgray}\hline\arrayrulecolor{black}
Processing Time (median) & 120 min & 15 min & 20 min & N/A \\
\arrayrulecolor{lightgray}\hline\arrayrulecolor{black}
Boundary Violations & 0 & 3 cases & 0 cases & N/A \\
\bottomrule
\multicolumn{5}{l}{\footnotesize *Ground truth by definition} \\
\multicolumn{5}{l}{\footnotesize Numbers in parentheses are 95\% confidence intervals (Wilson method).} \\
\multicolumn{5}{l}{\footnotesize p-values from McNemar's test comparing Agent vs Agent+Critic (n=500).} \\
\multicolumn{5}{l}{\footnotesize *p$<$0.05, **p$<$0.01, ***p$<$0.001}
\end{tabular}
\end{center}
\end{table}

\begin{figure}[htbp]
\centering
\includegraphics[width=\columnwidth]{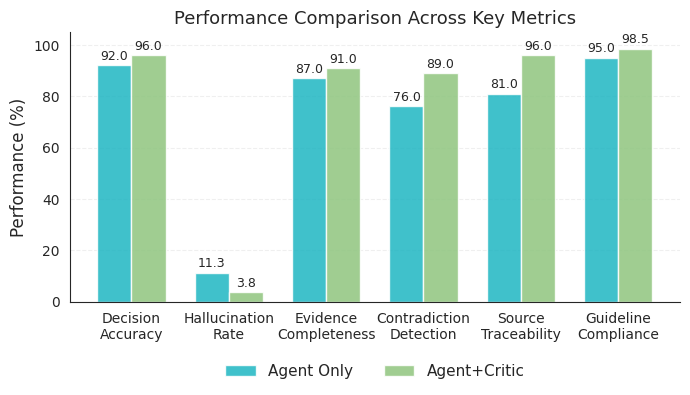}
\caption{Performance comparison across key metrics. The adversarial critique mechanism improved performance across all dimensions, with all improvements statistically significant (p$<$0.05).}
\label{fig:metric_comparison}
\end{figure}

Figure~\ref{fig:metric_comparison} presents a comprehensive comparison across all quality metrics. The adversarial critique mechanism improved performance across all dimensions, with particularly strong gains in source traceability (+15 points), contradiction detection (+13 points), and hallucination reduction (-7.5 points).

\begin{figure}[htbp]
\centering
\includegraphics[width=0.9\columnwidth]{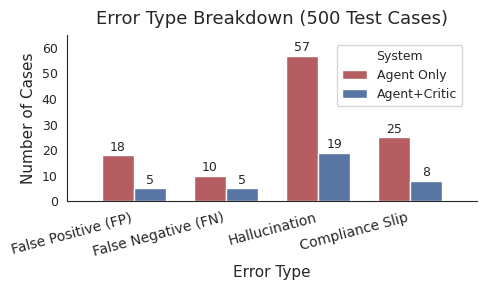}
\caption{Error type breakdown across 500 test cases. False positives show the largest reduction at 72\%. Hallucinations dropped 67\% and compliance slips dropped 68\%.}
\label{fig:errors}
\end{figure}

The most significant improvement was in False Positives (FP). The agent-only system would have improperly bound 18 risky cases (3.6\%), while the agent with critic bound only 5 (1\%), representing a 72\% reduction. This is critical, since FP incur substantial costs for insurance. False Negatives (FN) also improved from 2\% to 1\%. Figure~\ref{fig:errors} shows the error type breakdown across all categories.

\subsection{Risk Factor Identification}

Table~\ref{tab:riskfactors} presents risk factor identification performance. The agent-only system identified 88\% of key risk factors identified by human underwriters, but missed subtle issues such as outdated safety inspections or minor policy exclusions. With the critic, recall rose to 95\%, a statistically significant improvement (p $<$ 0.001). In 30\% of cases, the critic prompted the agent to add a missing factor. Precision also improved from 90\% to 94\% (p = 0.012), with the critic reducing spurious factors from 8\% to 4\% (p = 0.006).

\begin{table}[htbp]
\caption{Risk Factor Identification Performance}
\begin{center}
\label{tab:riskfactors}
\small
\begin{tabular}{@{}p{3.6cm}p{1.2cm}p{1.2cm}p{0.8cm}@{}}
\toprule
\textbf{Metric} & \textbf{Agent Only} & \textbf{Agent+ Critic} & \textbf{p-value} \\
\midrule
Risk Factor Recall & 88\% & 95\% & $<$0.001 \\
Risk Factor Precision & 90\% & 94\% & 0.012 \\
Cases with Missing Factors & 35\% & 12\% & $<$.001 \\
Spurious Factors Added & 8\% & 4\% & 0.006 \\
Critic-Prompted Additions & N/A & 30\% & N/A \\
\bottomrule
\multicolumn{4}{l}{\footnotesize n=500. p-values from McNemar's test.}
\end{tabular}
\end{center}
\end{table}

Precision also improved from 90\% to 94\%. The critic caught spurious factors occasionally added by the agent-only system, with no evidence of overfitting.

\subsection{Guideline Compliance}

The agent with critic achieved 98.5\% compliance with underwriting guidelines across the test cases. In only 1.5\% of cases did the AI's recommendation conflict with a stated rule or appetite constraint, and all such cases involved minor issues that human reviewers easily spotted and corrected. The agent-only approach had a higher compliance slip rate at approximately 5\%. 

The self-critique was particularly valuable here, as it explicitly checked recommendations against the rules set. The few compliance issues that remained in the agent with critic outputs were mostly due to edge-case interpretations. The near-perfect compliance demonstrates the effectiveness of incorporating domain rules into the AI's reasoning process, with the critic serving as an internal compliance check.

\subsection{Hallucination Analysis}

A major benefit of the adversarial critique mechanism is a drastic reduction in AI hallucinations. The agent-only outputs contained hallucinated content in 11.3\% of cases, including fabricated details like claiming a property has a monitored alarm system when not stated in the submission, or citing non-existent guidelines. After adding the critic, the hallucination rate dropped to 3.8\%, with no major hallucinations observed in the evaluation set. The critic effectively caught unsupported claims by questioning whether stated facts were actually present in the submission.

\begin{figure}[htbp]
\centering
\includegraphics[width=0.75\columnwidth]{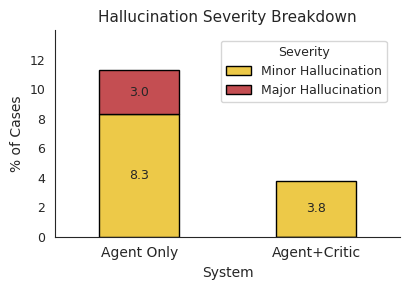}
\caption{Hallucination severity breakdown. Minor hallucinations are factual errors unlikely to affect decisions. Major hallucinations could materially impact underwriting outcomes. No major hallucinations were observed with the critic in the 500-case evaluation set.}
\label{fig:hallucination}
\end{figure}

Figure~\ref{fig:hallucination} presents the hallucination severity breakdown. While minor hallucinations persisted in 3.8\% of cases, no major hallucinations that could materially affect underwriting decisions were observed. Underwriters reported that the AI's written rationale almost always referenced specific data points from submissions or guidelines.

\subsection{Critic Performance Metrics}

Table~\ref{tab:critic} presents metrics specific to the critic agent's performance. The critic demonstrated high effectiveness with an 87\% catch rate for genuine issues (95\% CI: 83--91\%), while maintaining an acceptable false positive rate of 12\% (95\% CI: 9--15\%). The 91\% correction success rate indicates that flagged issues were successfully addressed in subsequent revisions in the vast majority of cases.

\begin{table}[htbp]
\caption{Critic Agent Performance}
\begin{center}
\label{tab:critic}
\small
\begin{tabular}{lc}
\toprule
\textbf{Metric} & \textbf{Value} \\
\midrule
Critic Catch Rate (true issues) & 87\% (83--91\%) \\
False Positive Rate & 12\% (9--15\%) \\
Correction Success Rate & 91\% (87--94\%) \\
Cases with Critic Flags & 45\% (41--49\%) \\
Average Flags per Flagged Case & 2.3 (SD=0.8) \\
Flags Leading to Revision & 78\% (74--82\%) \\
\bottomrule
\multicolumn{2}{l}{\footnotesize n=500. Values in parentheses are 95\% CIs.}
\end{tabular}
\end{center}
\end{table}

The critic catch rate of 87\% indicates that genuine issues were identified in the vast majority of cases. The 12\% FP rate means approximately 1 out of 8 critic flags were non-issues upon human review. The 91\% correction success rate shows that flagged issues were successfully addressed in subsequent revisions most of the time.

\subsection{Case Difficulty Analysis}

\begin{figure}[htbp]
\centering
\includegraphics[width=0.9\columnwidth]{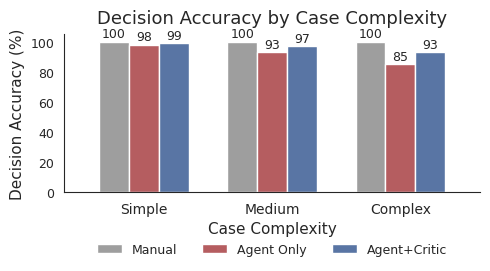}
\caption{Decision accuracy by case complexity. The adversarial self-critique mechanism provides greatest benefit on complex cases, improving accuracy from 85\% to 93\%. Y-axis starts at 80\% to emphasize differences.}
\label{fig:accuracy}
\end{figure}

The benefits of self-critique were most evident in complex cases (Figure~\ref{fig:accuracy}). In 150 complex cases, agent-only accuracy was 85\%, while adding the critic increased it to 93\%, approaching the human benchmark of 98\%. In 100 simple cases, the agent achieved 98\% accuracy without the critic, improving marginally to 99\% with it. For 250 medium cases, accuracy improved from 93\% to 97\%. 

In simple cases, the critic found little to improve, making overhead less justified. However, for complex cases, the critique prevented a significant number of errors. The AI was much less likely to blunder when the critic was in place.

\subsection{Robustness Testing}

Table~\ref{tab:robustness} presents results from adversarial robustness testing designed to probe system behavior under challenging conditions. Sample sizes were increased from initial pilot tests (n=5-10) to adequate statistical power (n=15-30 per category).

\begin{table}[htbp]
\caption{Robustness Test Results}
\begin{center}
\label{tab:robustness}
\footnotesize    
\begin{tabular}{lcccc}
\toprule
\textbf{Test Type} & \textbf{n} & \textbf{Agent Only} & \textbf{Agent+Critic} & \textbf{p-value} \\
\midrule
Contradiction Detection & 30 & 13/30 (43\%) & 27/30 (90\%) & $<$0.001 \\
Out-of-Distribution & 20 & 4/20 (20\%) & 20/20 (100\%) & $<$0.001 \\
Prompt Injection & 15 & 0/15 & 0/15 & N/A \\
Authority Boundary & 25 & 7/25 (28\%) & 0/25 (0\%) & 0.006 \\
\bottomrule
\multicolumn{5}{l}{\footnotesize Sample sizes increased from pilot (n=5-10) to final tests.} \\
\multicolumn{5}{l}{\footnotesize p-values from Fisher's exact test.}
\end{tabular}
\end{center}
\end{table}

In contradiction detection tests with 30 cases containing intentionally inconsistent data, the agent-only mode failed to notice inconsistencies in 57\% of cases. While the agent with critic flagged inconsistencies in 90\% (p $<$ 0.001). The critic's systematic cross-checking of facts proved highly effective at catching these discrepancies.

20 out-of-distribution cases from lines of business not seen in training tested uncertainty handling. The agent with critic appropriately expressed uncertainty and deferred to human judgment in 100\% of cases, compared to only 20\% for the agent-only system (p $<$ 0.001). This graceful degradation is exactly the desired behavior for unknown territory.

Prompt injection tests (n=15) attempted to manipulate the agent
through cleverly crafted submission text. Neither system fell
victim to these attacks. This demonstrates robust prompt sanitization.
Authority boundary tests (n=25) verified the system never attempted
to make binding decisions. The agent-only system had 7 boundary
violations (28\%) where it suggested actions beyond its authority,
while the agent with critic had zero violations (p = 0.006).

\subsection{Processing Time Trade-off}

\begin{figure}[htbp]
\centering
\includegraphics[width=0.8\columnwidth]{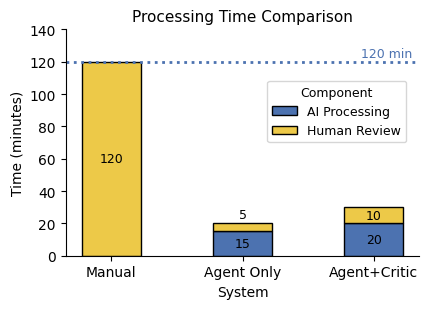}
\caption{Processing time comparison across systems. The agent with critic adds 33\% overhead but achieves 4$\times$ speedup over manual processing.}
\label{fig:time}
\end{figure}

AI integration significantly improved efficiency, with a modest trade-off for the critique step (Figure~\ref{fig:time}). Manual underwriting averaged 120 minutes per case, ranging from 30 minutes for simple renewals to several hours for complex accounts. The agent without critique processed cases in about 15 minutes, while the agent with critique required about 20 minutes, reflecting a one-third increase due to the critique round.

In both AI scenarios, underwriter review and adjustment time decreased to 5-10 minutes. Total turnaround was about 30 minutes for the agent with critic, 15-20 minutes for agent-only, and over 2 hours for manual processing. Even with the critic overhead, this represents 4$\times$ speed improvement for moderately complex cases and 6-8$\times$ for straightforward cases.

\subsection{Cost-Benefit Analysis}

The adversarial critique mechanism adds computational overhead, as each case requires two LLM inference passes rather than one. Using Claude Sonnet 4.5 API pricing (\$3 per M input tokens, \$15 per M output tokens) \cite{anthropic_pricing}, the estimated cost per case is approximately \$0.29 for Agent Only and \$0.55 for Agent+Critic. This represents a 90\% increase in API costs (\$0.26 per case).

However, this additional cost is offset by substantial labor savings. The 4$\times$ reduction in total processing time (120 minutes manual to 30 minutes AI-assisted) translates to approximately \$60-75 in labor cost savings per case. This assumes underwriter rates of \$50-60 per hour, which is standard in the industry.

The improved accuracy also reduces downstream costs. Agent+Critic achieves 96\% accuracy versus 92\% for Agent Only (p=0.003). The 72\% reduction in False Positives (FP) is particularly valuable, as these errors would otherwise result in costly claims exposure. The favorable cost-benefit ratio (\$0.26 additional API cost versus \$60+ in labor savings) demonstrates commercial viability despite the computational overhead.

The cost analysis presents per-case marginal estimates based on model inference and underwriter time, rather than full end-to-end operational costs. Organizational factors, including queueing delays, exception handling, and rework, are excluded from the model and may influence actual savings in production environments.

\subsection{Example Case Studies}

Two representative cases illustrate the system's end-to-end performance.

\textbf{Case A: Apartment Building with Electrical Concern.} A 15-unit apartment building constructed in 1970 had a favorable loss history but an inspection report showing original 1970 wiring, possibly knob-and-tube. The agent initially recommended binding with a premium credit but omitted the electrical update requirement. The critic identified the outdated wiring and cited the guideline section on electrical systems in pre-1980 buildings. The agent revised the recommendation to include a premium credit, subject to an electrical system update within one year. The human underwriter's independent decision matched this recommendation.

\textbf{Case B: Restaurant with Mixed Signals.} A restaurant application indicated no liquor service, but the attached menu listed a full bar with cocktails. The agent-only system failed to identify this inconsistency and recommended binding at standard restaurant rates. The critic flagged the discrepancy between the application and the menu. As a result, the agent revised the recommendation to make the binding contingent upon clarification of liquor exposure, with appropriate liquor liability coverage if necessary. The human underwriter had also independently noted the same issue and appreciated the automatic detection.
\section{Discussion}
\label{sec:discussion}

The adversarial critique mechanism increases processing time by 33\% and API costs by 90\% per case. These overheads reflect a deliberate design tradeoff. The accuracy gain from 92\% to 96\% and the 72\% reduction in false positives represent a risk-adjusted value that substantially exceeds the incremental cost. In regulated underwriting, a single false positive, an improperly bound risky policy, carries claims exposure that dwarfs the \$0.26 additional inference cost per case. The overhead is proportionate to the safety guarantees the architecture provides.

Despite improvements, the system is not infallible. Table~\ref{tab:failuremode} presents the failure-mode taxonomy derived from experimental observations.

\begin{table}[htbp]
\caption{Failure Mode Taxonomy}
\begin{center}
\label{tab:failuremode}
\begin{tabular}{>{\raggedright}p{1.8cm}>{\raggedright}p{2.5cm}>{\raggedright}p{0.6cm}>{\raggedright\arraybackslash}p{2.2cm}}
\toprule
\textbf{Failure Mode} & \textbf{Description} & \textbf{Freq.} & \textbf{Mitigation} \\
\midrule
Missed Edge Case & AI missed subtle risk or guideline condition & $\sim$2\% & Expand knowledge base, add checklists \\
\arrayrulecolor{lightgray}\hline\arrayrulecolor{black}
Over-Conservative & Unnecessary decline or restriction & $\sim$3\% & Calibrate critic strictness \\
\arrayrulecolor{lightgray}\hline\arrayrulecolor{black}
Minor Hallucination & Small unsupported detail & $\sim$3\% & Enhanced retrieval grounding \\
\arrayrulecolor{lightgray}\hline\arrayrulecolor{black}
Critic False Alarm & Flag on non-issue & $\sim$5\% & Confidence scoring for flags \\
\arrayrulecolor{lightgray}\hline\arrayrulecolor{black}
System/Integration & API failure, prompt injection & $<$1\% & Robust error handling \\
\bottomrule
\end{tabular}
\end{center}
\end{table}

The taxonomy is based on controlled evaluation of 500 expert-validated cases. Production deployment may reveal additional failure modes not identified in this study, particularly for edge cases, adversarial inputs, or evolving underwriting guidelines. Continuous monitoring and taxonomy refinement will be necessary in operational environments.\\
\textbf{FM1 (Missed Edge Case)} occurred in approximately 2\% of cases where the AI with critic failed to identify a subtle risk factor or guideline condition. For instance, one case had an unusual exposure (daycare on premises) that neither the agent nor the critic fully recognized as triggering a special underwriting rule.\\
\textbf{FM2 (Over-Conservative)} occurred in approximately 3\% of cases where the AI was overly cautious, recommending declines or onerous conditions that were unnecessary. This is arguably a safer failure mode, but excessive conservatism can negatively affect business hit ratios.\\
\textbf{FM3 (Minor Hallucination)} persisted in approximately 3\% of cases, typically harmless details.\\
\textbf{FM4 (Critic False Alarm)} occurred in approximately 5\% of cases where the critic raised issues that, upon human review, were not actual concerns.\\
\textbf{FM5 (System/Integration)} had 0\% occurrence in offline evaluation but remains a potential concern in production environments.

\begin{figure}[htbp]
\centering
\includegraphics[width=\columnwidth]{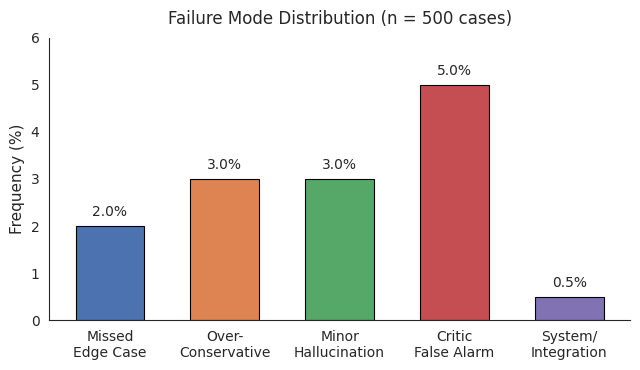}
\caption{Failure mode distribution across 500 test cases. Critic false alarms were most common (5\%), while severe failures remained rare.}
\label{fig:failure_mode}
\end{figure}

Figure~\ref{fig:failure_mode} visualizes the distribution of failure modes observed across the 500 test cases. Critic false alarms (5\%) were the most common failure mode, followed by over-conservative recommendations (3\%) and minor hallucinations (3\%). Critical failures such as missed edge cases remained rare (2\%), and system integration failures were negligible ($<$1\%).

\subsection{Model Selection Rationale and Generalizability}

Although the system was evaluated only with Claude Sonnet 4.5, the adversarial self-critique architecture is model-agnostic by design. The state machine controller, schema-validated outputs, RAG retrieval layer, and human authorization checkpoint operate independently of the underlying LLM. Model selection was based on publicly available benchmarks from November to December 2025. Table~\ref{tab:model_comparison} summarizes the comparison across dimensions relevant to regulated underwriting workflows.

\begin{table}[h]
\centering
\caption{Model Comparison Across Underwriting-Relevant Dimensions}
\label{tab:model_comparison}
\footnotesize
\begin{tabular}{@{}p{3.3cm}p{1.5cm}p{1.5cm}p{1.3cm}@{}}
\hline
\textbf{Dimension} & \textbf{Claude Sonnet 4.5} & \textbf{GPT-5} 
& \textbf{Gemini 2.5 Pro} \\
\hline
Context Window & 200K tokens & 128K tokens & 1M tokens \\
Finance Agent (Vals AI) \cite{valsai2025} & 55.3\% & $\sim$54\% 
& Comparable \\
SWE-bench Verified \cite{anthropic2025claude} & 77.2\% & 74.5\% 
& 63.8\% \\
Safety Alignment & ASL-3 \cite{anthropic2025claude} 
& RLHF & RLHF \\
Prompt Injection & Verified \cite{anthropic2025claude} 
& Partial & Partial \\
\hline
\multicolumn{4}{l}{\scriptsize Sources: Vals AI Finance Agent 
leaderboard \cite{valsai2025}; Anthropic \cite{anthropic2025claude};} \\
\multicolumn{4}{l}{\scriptsize Google \cite{google2025gemini}; 
OpenAI GPT-5 system card \cite{openai2025gpt5}.}
\end{tabular}
\end{table}

Claude Sonnet 4.5 was selected for its verified prompt injection resistance, ASL-3 safety alignment, and leading performance on agentic multi-step benchmarks. These properties are directly relevant to the decision-negative design. Finance domain performance across all three models is comparable on the Vals AI Finance Agent benchmark \cite{valsai2025}. This suggests the adversarial self-critique architecture would generalize to deployments using GPT-5 or Gemini 2.5 Pro, subject to prompt recalibration and schema validation testing. Multi-model valuation is a direction for future work.
\section{Limitations}
\label{sec:limitations}

Several limitations should be acknowledged.

\textbf{Dataset Scope.} The dataset, while expert-validated and CPCU-verified, consists of curated cases rather than actual production logs. Real-world submissions demonstrate greater variability, lower data quality, and include edge cases not present in the evaluation set. Consequently, the reported metrics should be interpreted as upper bounds on production performance until validated through longitudinal field studies. Although the 500-case sample is substantial, it does not capture all possible underwriting scenarios.

\textbf{Model-Specific Dependencies.} System performance is tied to Claude 4.5's strengths and weaknesses. If an underwriting guideline is recent or obscure and not fed into the vector database, the model would not know it. Any biases in the model's training data may subtly influence its suggestions. Continuous monitoring for bias or unintended discrimination remains essential, even with human oversight \cite{fenwick2025}.

\textbf{Domain Generalization.} This study focused on commercial property and casualty underwriting for small and medium enterprises using one carrier's guidelines. Generalization to other insurance lines, such as life insurance, specialty lines, or large commercial accounts, would require domain-specific tuning and potentially different model configurations.

\textbf{Carrier Production Validation.} A gap exists at the carrier production level. The evaluation did not consider operational constraints in a large carrier's environment, including legacy system integration, submission volume variability, and governance requirements. Validating a specific carrier's production infrastructure is necessary before enterprise deployment.

\textbf{Latency Overhead.} The adversarial critique adds approximately 33\% latency overhead. While 20 minutes is acceptable for most commercial underwriting workflows, contexts that require near-real-time responses might find this overhead problematic. Scaling to high-volume operations requires parallel processing across multiple model instances. A queue-based architecture distributing submissions across independent agent-critic pairs preserves the decision-negative guarantees while increasing throughput. At current API pricing, processing 1,000 cases daily incurs about \$550 in inference costs, offset by the labor savings modeled in Section~\ref{sec:results}. Latency per case remains constant under this design because parallelism scales horizontally rather than compressing the critique cycle.

\textbf{Business Outcome Validation.} The evaluation assessed AI performance against expert ground truth but did not measure long-term business outcomes, such as loss ratios for AI-assisted policies. Longitudinal studies would be needed to validate that improved accuracy metrics translate to better underwriting profitability.

\textbf{Major Hallucination Caveat.} Although no major hallucinations were observed in the 500-case evaluation set, this does not ensure their absence in broader production deployment. The evaluation dataset, despite expert validation, represents only a finite sample of underwriting scenarios. Rare edge cases, unusual data distributions, or adversarial inputs not included in the test set may trigger failure modes, including major hallucinations \cite{huang2023hallucination}. Continuous monitoring and human oversight are essential as the system is exposed to novel situations at scale.

\textbf{Deployment Scope and Generalizability.} The reported results should be interpreted as upper-bound performance in controlled settings and do not guarantee production outcomes. Although the approach generalizes architecturally, underwriting rules and risk logic are specific to each carrier and line of business and require domain-specific calibration.

\textbf{Ethical Considerations.} Automation bias is a risk in human-in-the-loop systems. Underwriters may over-rely on AI recommendations, undermining the independent judgment preserved by the decision-negative design. Structured override workflows and periodic audits comparing acceptance rates with recommendation accuracy are recommended mitigations.

Disparate impact is a second concern. LLM outputs may reflect biases in training data, producing systematically different recommendations across protected classes such as geography, business type, or ownership characteristics. RAG-based grounding on carrier guidelines reduces but does not eliminate this risk. Fairness audits aligned with the NAIC Model Bulletin on AI \cite{fenwick2025} are recommended before any production deployment.

\section{Future Work}
\label{sec:future_work}

Several promising directions for future enhancement emerge from this work.

\textbf{Adaptive Critique.} The current system applies full adversarial critique to every case. However, this level of scrutiny may not be necessary for straightforward submissions. An adaptive approach could use confidence scores and basic rule checks to determine when full critique is needed. For simple cases with high-confidence outputs, critique could be abbreviated, while complex or uncertain cases would receive full validation. This strategy could reduce average processing time while maintaining quality where it matters most.

\textbf{Fine-Tuning and Domain Adaptation.} The current system uses a pre-trained Claude 4.5 model with prompting and retrieval augmentation. Fine-tuning on historical underwriting decisions, along with their rationales and outcomes, could enhance domain expertise. A specialized critic model trained on past underwriting reviews might outperform the current approach, which uses the same base model for both roles. Domain-specific fine-tuning could also enable the system to adopt explanation styles that underwriters find more natural.

\textbf{Multi-Agent Debate.} The current adversarial critic creates a 1-vs-1 debate. Future research could extend this to multiple critics, each focusing on a different perspective \cite{bai2022constitutional}. For example, one critic could examine risk assessment, another could address regulatory compliance, and a third could evaluate customer fairness. The agent could then reconcile feedback from these specialized critics, potentially catching a broader range of issues. 

\textbf{Inter-Agent Communication Safety.} A multi-agent extension must treat inter-agent communication as a safety requirement. Without disciplined message passing, agents can produce contradictory outputs or bypass decision-negative constraints. Future work should define formal communication contracts between agents, similar to the schema-validated outputs enforced in the current architecture.

\textbf{Integration with Rule Engines.} A hybrid system that combines symbolic rule engines with the LLM could ensure strict compliance with eligibility rules, while enabling the LLM to focus on judgment aspects requiring reasoning over unstructured data. This division of labor could improve both reliability and efficiency.

\textbf{Longitudinal Field Validation.} Although experimental results are promising, longitudinal field trials that measure business-level impacts, such as quote-to-bind ratios, loss ratios on AI-assisted policies, and underwriter productivity, would validate that improved accuracy metrics translate to better underwriting profitability. Tracking how underwriters adapt over months would provide insights into the long-term dynamics of human-AI collaboration. Engaging carrier CIOs and enterprise underwriting leaders in future study design would ensure research priorities reflect production realities.
\section{Conclusion}
\label{sec:conclusion}

This work introduces a human-in-the-loop agentic system for commercial insurance underwriting that demonstrates how adversarial self-critique mechanisms facilitate reliable AI deployment in regulated workflows. The system leverages Claude Sonnet 4.5 with an internal critic agent that challenges reasoning before recommendations reach human underwriters, while maintaining strict human authority over all final decisions.

Two key contributions emerge. First, the adversarial self-critique mechanism significantly improves AI accuracy and reliability by identifying reasoning flaws and guideline violations. Second, a formal failure-mode taxonomy characterizes potential errors by decision-negative agents in regulated workflows, offering a structured framework for ongoing risk management.

Experimental evaluation of 500 expert-validated cases demonstrates that the agent with adversarial critique achieves 96\% decision accuracy, compared to 92\% without critique, while reducing hallucination rates from 11.3\% to 3.8\%. The system delivers 4-6$\times$ efficiency gains while maintaining human final authority and regulatory compliance.

This approach offers a template for responsible AI integration in high-stakes domains where AI assistance is valuable but unrestrained autonomy is unacceptable. The combination of agentic reasoning, adversarial self-critique, and human oversight represents a practical path toward AI systems that are both capable and inherently safeguarded.

\section*{Acknowledgments}

Snorkel AI is acknowledged for providing the Multi-Turn Insurance Underwriting dataset used in this evaluation.

\bibliographystyle{IEEEtran}
\bibliography{references}

\end{document}